\documentclass[lettersize,journal]{IEEEtran}
\usepackage{amsmath,amsfonts}
\usepackage{algorithmic}
\usepackage{algorithm}
\usepackage{array}
\usepackage[caption=false,font=normalsize,labelfont=sf,textfont=sf]{subfig}
\usepackage{textcomp}
\usepackage{stfloats}
\usepackage{url}
\usepackage{verbatim}
\usepackage{graphicx}
\usepackage{cite}
\usepackage{color}
\usepackage{graphicx}
\usepackage{amsmath}
\usepackage{amssymb}
\usepackage{booktabs}
\usepackage{bm}
\usepackage{diagbox}
\usepackage{subfig}
\usepackage[T1]{fontenc}
\usepackage[table]{xcolor}
\begin{document}

\title{Controllable Relation Disentanglement for Few-Shot \\Class-Incremental Learning}
\author{Yuan Zhou, Richang Hong, Yanrong Guo, Lin Liu, Shijie Hao, Hanwang Zhang
\thanks{Yuan Zhou, Richang Hong, Yanrong Guo, and Shijie Hao are with Key Laboratory of Knowledge Engineering with Big Data (Hefei University of Technology), Ministry of Education, and School of Computer Science and Information Engineering, Hefei University of Technology, Hefei 230009, China (e-mail: zhouyuan888888@gmail.com.cn, hongrc.hfut@gmail.com, yrguo@hfut.edu.cn, hfut.hsj@gmail.com). Richang Hong is the corresponding author. Lin liu is with Department of Electronic Engineering and Information Science, University of Science and Technology of China, Hefei, 230027, China (e-mail: ll0825@mail.ustc.edu.cn). Hanwang Zhang is with the School of Computer Science and Engineering, Nanyang Technological University, Singapore, 639798 (e-mail: hanwangzhang@ntu.edu.sg).}}


\markboth{Journal of \LaTeX\ Class Files,~Vol.~14, No.~8, August~2021}%
{Shell \MakeLowercase{\textit{et al.}}: A Sample Article Using IEEEtran.cls for IEEE Journals}


\maketitle

\begin{abstract}
    In this paper, we propose to tackle Few-Shot Class-Incremental Learning (FSCIL) from a new perspective, \emph{i.e., relation disentanglement}, which means enhancing FSCIL via disentangling spurious relation between categories. The challenge of disentangling spurious correlations lies in the poor controllability of FSCIL. On one hand, an FSCIL model is required to be trained in an incremental manner and thus it is very hard to directly control relationships between categories of different sessions. On the other hand, training samples per novel category are only in the few-shot setting, which increases the difficulty of alleviating spurious relation issues as well. To overcome this challenge, in this paper, we propose a new simple-yet-effective method, called \textbf{C}on\textbf{T}rollable \textbf{R}elation-disentang\textbf{L}ed \textbf{F}ew-\textbf{S}hot \textbf{C}lass-\textbf{I}ncremental \textbf{L}earning (CTRL-FSCIL). Specifically, during the base session, we propose to anchor base category embeddings in feature space and construct disentanglement proxies to bridge gaps between the learning for category representations in different sessions, thereby making category relation controllable. During incremental learning, the parameters of the backbone network are frozen in order to relieve the negative impact of data scarcity. Moreover, a disentanglement loss is designed to effectively guide a relation disentanglement controller to disentangle spurious correlations between the embeddings encoded by the backbone. In this way, the spurious correlation issue in FSCIL can be suppressed. Extensive experiments on CIFAR-100, mini-ImageNet, and CUB-200 datasets demonstrate the effectiveness of our CTRL-FSCIL method.
 
\end{abstract}


\section{Introduction}

\IEEEPARstart{D}{eep} learning has achieved remarkable success in recent years due to the availability of large-scale training data. However, the performance of current deep learning models is still far from satisfactory in the low-data scenario, especially in the incremental learning manner. Overcoming this challenge is crucial for further improving the applicability of these models in real-world applications. To this end, a new Few-Shot Class-Incremental Learning (FSCIL) task has been introduced by Tao et al. \cite{tao2020few}, which focuses on helping models to learn novel unseen categories according to only a few training data while trying to maintain old knowledge acquired from previous categories.

Since the proposal of FSCIL, it has attracted much attention from the community due to its huge value in practical applications. The FSCIL task is more challenging than Class-Incremental Learning (CIL) \cite{hinton2015distilling,rebuffi2017icarl,isele2018selective} and Few-Shot Learning (FSL) \cite{vinyals2016matching,snell2017prototypical} as it considers the negative impact of incremental learning and data scarcity problems at the same time. Current FSCIL methods are mainly based on knowledge distillation \cite{cui2023uncertainty,cheraghian2021semantic,zhao2023few,dong2021few}, margin-based regularization \cite{peng2022few,zou2022margin}, distribution rectification \cite{liu2023learnable,cheraghian2021synthesized,akyurek2021subspace}, and embedding reservation \cite{zhou2022forward,song2023learning}. Knowledge-distillation-based methods \cite{cui2023uncertainty,cheraghian2021semantic,zhao2023few,dong2021few} aim at ensuring the effectiveness of models on previously learned old categories, so as to suppress catastrophic-forgetting issues in incremental learning as much as possible. Differently, margin-based regularization \cite{peng2022few,zou2022margin} and embedding reversion \cite{zhou2022forward,song2023learning} approaches concentrate on guiding a model to learn a compact representation space, therefore improving the transferability of the model and accordingly making it more effective to tackle subsequently encountered novel class tasks \cite{peng2022few,zou2022margin,zhou2022forward,song2023learning}. Approaches based on distribution rectification \cite{liu2023learnable,cheraghian2021synthesized,akyurek2021subspace} focus on strengthening a model by fully considering the distribution information of given data and effectively boosting the performance on both base and novel categories. 

\begin{figure}[t!]
\centering
\includegraphics[height=4cm]{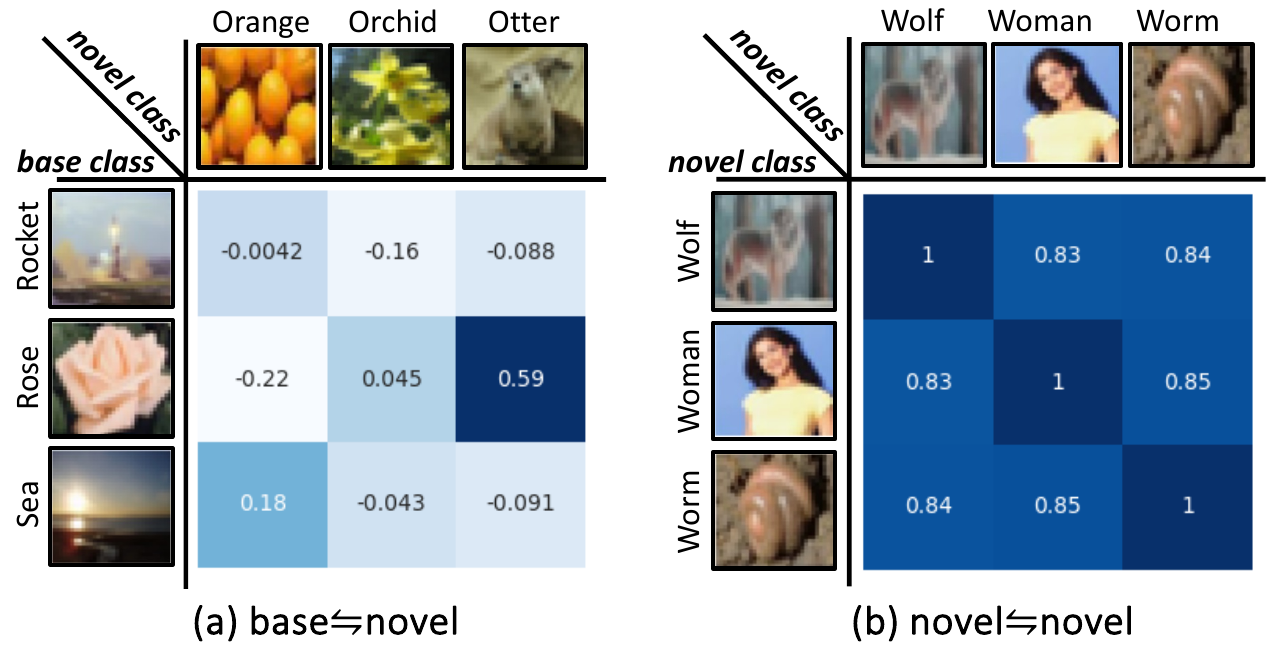}
\caption{Typical examples for spurious relation issues in few-shot class-incremental learning, which are obtained from the last incremental session of the CIFAR-100 dataset. The values in the matrixes indicate correlations between the learned global embeddings of categories. Also, ``base$\leftrightharpoons$novel'' represents spurious relation issues between base and novel categories, while ``novel$\leftrightharpoons$novel'' indicates spurious relation issues between novel categories.}
\label{fig:a}
\end{figure}

Despite the progress achieved in this task, spurious relation issues are usually ignored by current FSCIL approaches. Specifically, on one hand, the training samples of base categories are sufficient, and accessible in the base session at the same time. As a result, a model can be easily trained to properly perceive relationships between base categories. However, on the other hand,  when the model is extended to unseen categories in incremental sessions, the joint influence of incremental learning and data scarcity induces severe spurious relation issues between base and novel, or novel and novel categories. As examplified in Fig. \ref{fig:a}, the model fails to clearly distinguish the novel category ``Rose'' and the base category ``Otter'' during incremental learning, and wrongly assigns a high correlation value for them. In fact, the category ``Rose'' and the category ``Otter'' have large discrepancies whatever in appearance and semantic space. Similarly, the model is also biased towards the hypothesis that the novel category ``Worm'' is highly correlated to the novel categories ``Wolf'' and ``Woman''. These spurious relation issues largely intensify inter-class interference of encountered categories and therefore prevent FSCIL models from accurately predicting the labels of samples.

Based on the above observations, in this paper, we propose to tackle the FSCIL task from a new perspective, \emph{i.e., relation disentanglement}, which means solving few-shot class-incremental learning via disentangling spurious correlations between categories. However, directly disentangling spurious relation between categories is difficult due to the poor controllability of training an FSCIL model. On one hand, the FSCIL model is required to be trained in an incremental manner. Thus, the samples of seen categories and subsequently encountered new ones are inaccessible to participate in training at the same time. This asynchrony enlarges the gaps between the learning for the representations of encountered categories and makes it difficult to guide the model to perceive category relation properly. On the other hand, the training samples of novel classes are only in the few-shot setting and thus data scarcity increases the difficulty of overcoming spurious relation issues as well.

To overcome these challenges, a new simple-yet-effective method, called \textbf{C}on\textbf{T}rollable \textbf{R}elation-disentang\textbf{L}ed \textbf{F}ew-\textbf{S}hot \textbf{C}lass-\textbf{I}ncremental \textbf{L}earning (CTRL-FSCIL), is proposed in this paper, which can effectively control relationships between categories. As shown in Fig.~\ref{fig:c}, our CTRL-FSCIL consists of two phases, namely \emph{controllable proxy learning} and \emph{relation-disentanglement-guided adaptation}, which are designed for base and incremental sessions respectively. Specifically, during the first phase, an orthogonal proxy anchoring strategy is utilized to anchor base class embeddings in feature space via fully considering the guidance from orthogonal proxies, thereby making base class embeddings controllable while suppressing inter-class interference of base categories. Meanwhile, a disentanglement proxy discriminability boosting strategy is adopted to build disentanglement proxies for subsequent novel categories, so as to bridge gaps between different sessions. Disentanglement proxies aim to ensure that the representations of novel categories have maximized discriminability as well as minimized correlations to base class features. In the second phase, the parameters of the backbone network are frozen in order to relieve the negative impact of data scarcity. Moreover, a relation disentanglement strategy is employed, where a disentanglement loss is designed to incrementally guide a relation disentanglement controller to rectify correlations of categories. In this way, the spurious relation issue in FSCIL can be alleviated.

All-in-all, the contributions of this paper can be concluded below:

\begin{itemize}
    \item[$\circ$] In this paper, we propose to tackle FSCIL from a new perspective, \emph{i.e., relation disentanglement}, which means solving the FSCIL task via disentangling spurious relation between categories. Accordingly, a simple-yet-effective method \textbf{C}on\textbf{T}rollable \textbf{R}elation-disentang\textbf{L}ed \textbf{F}ew-\textbf{S}hot \textbf{C}lass-\textbf{I}ncremental \textbf{L}earning (CTRL-FSCIL) is designed.
    \item[$\circ$] In the base session, we propose an orthogonal proxy anchoring strategy to anchor base class embeddings in feature space as well as a disentanglement proxy discriminability boosting strategy to build disentanglement proxies, so as to bridge gaps between sessions and effectively guide relation disentanglement during incremental learning.

    \item[$\circ$] Additionally, a relation disentanglement strategy is introduced in incremental sessions, which rectifies category relation by considering the guidance of disentanglement proxies. In this way, the spurious correlation issue can be suppressed.
    
    \item[$\circ$] Extensive experiments are conducted on CIFAR-100, mini-ImageNet, and CUB-200 datasets. Our method consistently shows competitive performance, \emph{e.g.}, achieving 66.84 \%, 68.97 \%, and 67.66~\% mean accuracy on these datasets respectively.
    \end{itemize}

\section{Related Work}
We review research that is relevant to our work in this section, including class-incremental learning, few-shot learning, and few-shot class-incremental learning, which are introduced in Section-II-A, Section-II-B, and Section-II-C respectively.

\subsection{Class-Incremental Learning}
Class-incremental learning (CIL) aims to continuously adapt a model to novel unseen categories while maintaining the knowledge previously learned from old ones. Catastrophic-forgetting issues have a large impact on the performance of CIL models. To overcome this challenge, data replay \cite{isele2018selective,aljundi2019gradient,rolnick2019experience} and knowledge distillation \cite{hinton2015distilling,rebuffi2017icarl,li2017learning,wu2019large} strategies have been widely used by current approaches, which are beneficial for further improving the consistency between old and incrementally updated new models. Also, \cite{liu2021rmm} enhanced the data replay strategy with a dynamic memory allocation mechanism, so as to avoid sub-optimal memory management problems. \cite{kang2022class} designed an adaptive-consolidation-based knowledge distillation strategy which restricts the drift of critical features though fully considering the relationships between distribution shift and loss change. Differently, \cite{douillard2020podnet} introduced a pooled output distillation loss, where spatial information is fully leveraged to help achieve a better trade-off between old knowledge maintenance and new category adaptation. Aiming to overcome a stability-plasticity dilemma, \cite{liu2021adaptive} proposed to learn stability and plasticity at each residual level via using stable and plastic blocks respectively, while \cite{yan2021dynamically} designed a dynamical expansion strategy to decouple the learning of feature representations and classifier weights. \cite{wang2022foster} introduced the concept of gradient boosting to CIL, and further improved the adaptiveness of the model in modeling novel categories. As CIL models require sufficient training instances, they are unsuitable to be applied to the scenario that training data are scarce.

\subsection{Few-Shot Learning}
Few-Shot Learning (FSL) aims to help a model to quickly learn novel visual concepts according to only a few labeled samples, thereby further reducing expenses cost on data preparation. An intuitive way to solve this task is using data augmentation strategies, such as data synthesis \cite{wang2018low,zhang2019few,li2020adversarial} and data retrieval \cite{wu2018exploit,douze2018low}. However, the effectiveness of these data augmentation strategies are usually limited by domain gaps \cite{wang2020generalizing}. Currently, metric-learning-based methods have dominated FSL due to their high effectiveness and flexibility \cite{vinyals2016matching,snell2017prototypical}. For these methods, semantic guidance and class-wise relationships were often utilized to further enhance metric classification \cite{li2019few,wang2020cooperative,li2020boosting,wang2020large,pahde2021multimodal,zhou2022hierarchical}. Augmenting FSL models with key-to-value memory is another hotspot in FSL \cite{santoro2016meta,kaiser2017learning, ramalho2019adaptive,cai2018memory}. For example, a key-to-value memory module was designed in \cite{santoro2016meta} to explicitly store the information of support samples in a more effective way. \cite{kaiser2017learning,ramalho2019adaptive} pointed out that it is necessary to suppress the redundancy of memory by only remembering rare events, which is beneficial for further improving classification accuracy. Differently, \cite{cai2018memory} utilized stored memory information to generate dynamical convolution kernels, so making it more effective to encode the embeddings of query samples. Recent research \cite{lin2023supervised,shirekar2023self} revealed that the use of the pre-training strategy is also a promising tendency for the FSL task. In general, current FSL models can exhibit excellent performance on novel category episodes. But they cannot distinguish novel and old categories jointly due to the failure of old knowledge maintenance.

\subsection{Few-Shot Class-Incremental Learning}
Few-Shot Class-Incremental Learning (FSCIL) can be seen as a combination of CIL and FSL tasks. It aims at helping models to learn novel categories according to only a few training instances while retaining old knowledge as much as possible. Aiming to better maintain old knowledge, relation, semantic, or uncertainty knowledge distillation was employed in \cite{cui2023uncertainty,cheraghian2021semantic,zhao2023few,dong2021few}, while \cite{liu2022few} introduced an entropy-regularized data replay strategy. In \cite{zhou2022forward,song2023learning}, forward-compatible FSCIL models are built to reserve embedding space for upcoming novel categories via using virtual prototypes. Different from \cite{zhou2022forward,song2023learning}, margin-based penalties were adopted in \cite{peng2022few,zou2022margin} to improve inter-class discriminability. In particular, \cite{peng2022few} pointed out that MixUp \cite{zhang2017mixup} is beneficial for further improving the quality of feature embeddings, while \cite{zou2022margin} proposed a negative-margin-based training strategy to effectively capture common embeddings shared by categories. \cite{chi2022metafscil} proposed a meta-learning-based FSCIL approach, where manually-built pseudo incremental episodes can eliminate the misalignment between training and testing stages. Also, the distribution information of given data was fully considered in \cite{liu2023learnable,cheraghian2021synthesized} to facilitate the categorization of query samples, and \cite{akyurek2021subspace} designed a subspace regularization scheme to limit the drift of novel class weights. Our CTRL-FSCIL is obviously different from the above methods, which aims to tackle FSCIL from a new perspective, \emph{i.e.} disentangling spurious relation between categories. 

\begin{figure*}[t!]
\centering
\includegraphics[height=10.2cm]{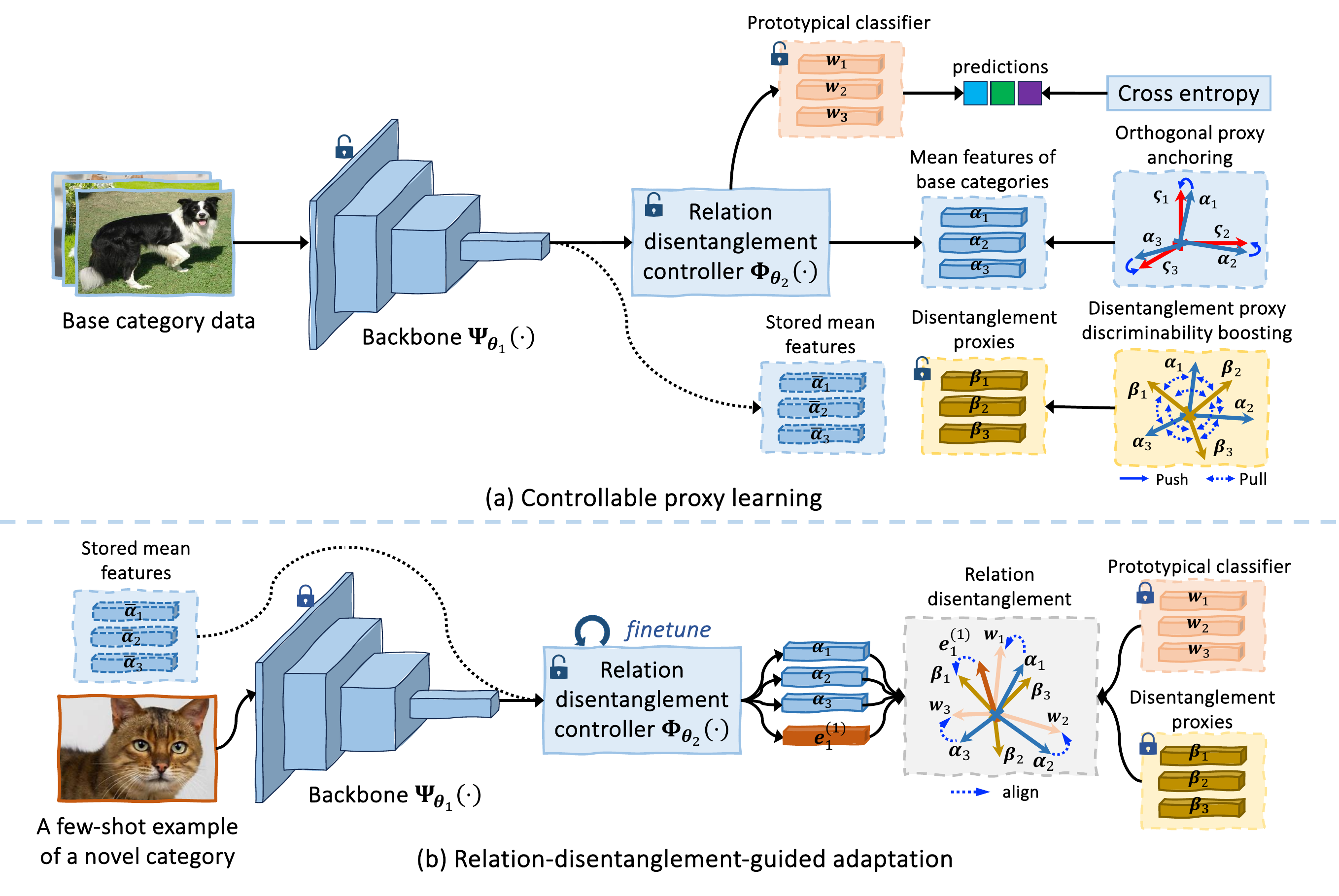}
\caption{An illustration of our \textbf{C}on\textbf{T}rollable \textbf{R}elation-disentang\textbf{L}ed \textbf{F}ew-\textbf{S}hot \textbf{C}lass-\textbf{I}ncremental \textbf{L}earning (CTRL-FSCIL) method, which is examplified in an incremental session.}
\label{fig:c}
\end{figure*}

\section{Methodology}
In this section, we first introduce the defination of FSCIL in Section III-A, and then elaborate the proposed CTRL-FSCIL in Section III-B. 

\subsection{Task Defination}
The FSCIL task aims to sequentially expand the model to novel categories according to only a small number of labeled training data given in each incremental session. In this work, the data given in the $t$-$th$ session are represented as $\bm{D}^{(t)}\in \{\bm{D}^{(1)}, \dots, \bm{D}^{(T)}\}$ and the label space of the dataset $\bm{D}^{(t)}$ is denoted as $\bm{C}^{(t)}\in \{\bm{C}^{(1)}, \dots, \bm{C}^{(T)}\}$. In incremental sessions, datasets obey the standard few-shot setting, generally termed as ``$|\bm{C}^{(t)}|$-way $K$-shot'', which means that the dataset $\bm{D}^{(t)}$ totally contains $|\bm{C}^{(t)}|$ categories and each category consists of $K$ training samples. In current FSCIL methods, a base dataset $\bm{D}^{(0)}$ is generally adopted to initialize the parameters of the FSCIL model. It contains significantly more categories and each category contains sufficient training instances, \emph{i.e.}, obeying the condition $|\bm{C}^{(0)}|\gg |\bm{C}^{(t)}|$ and $|\bm{D}^{(0)}|\gg |\bm{D}^{(t)}|$ for $t> 0$. Also, datasets from different sessions have a disjoint label space, which means $\bm{C}^{(t)}\cap \bm{C}^{(t')}=\emptyset$ for $\forall t'\ne t$. During the training in the $t$-$th$ session, only the dataset $\bm{D}^{(t)}$ can be accessed and the images from other sessions are inaccessible.

\subsection{Controllable Relation-Disentangled Few-Shot Class-Incremental Learning}
The framework of our \textbf{C}on\textbf{T}rollable \textbf{R}elation-disentang\textbf{L}ed \textbf{F}ew-\textbf{S}hot \textbf{C}lass-\textbf{I}ncremental \textbf{L}earning (CTRL-FSCIL) approach is given in Fig \ref{fig:c}. As can be seen from the figure, CTRL-FSCIL consists of two phases, which are \emph{controllable proxy learning} and \emph{relation-disentanglement-guided adaptation}. These two phases are designed for base and incremental sessions respectively. In the following, we introduce these two phases in detail.

\subsubsection{Controllable proxy learning}
In the first phase, we first use a backbone network $\bm{\Psi}_{\bm{\theta}_1}(\cdot)$ to extract the embeddings of inputs
\begin{equation}
\bm{h}^{(0)}_n = \bm{\Psi}_{\bm{\theta}_1}(\bm{X}^{(0)}_n)\text{,}
\label{eq:1}
\end{equation}

\noindent where $\bm{X}^{(0)}_n\in\{\bm{X}^{(0)}_{n}\}^{N_b}_{n=1}$ denotes one of a mini-batch of images sampled from the base dataset $\bm{D}^{(0)}$ and $\bm{h}^{(0)}_n\in\mathbb{R}^{1\times 1\times d}$ represents the extracted embeddings of the image $\bm{X}^{(0)}_n$. Additionally, for brevity, the labels of the images $\{\bm{X}^{(0)}_{n}\}^{N_b}_{n=1}$ are denoted as $\{y^{(0)}_{n}\}^{N_b}_{n=1}$. Then, a relation disentanglement controller $\bm{\Phi}_{\bm{\theta}_2}(\cdot)$ is further used to  process the features $\{\bm{h}^{(0)}_{n}\}^{N_b}_{n=1}$ and thereby generate the embeddings $\{\bm{e}^{(0)}_{n}\in\mathbb{R}^{1\times 1\times d'}\}^{N_b}_{n=1}$
\begin{equation}
\bm{e}^{(0)}_n = \bm{\Phi}_{\bm{\theta}_2}(\bm{h}^{(0)}_n)\text{.}
\label{eq:2}
\end{equation}

\noindent The backbone network $\bm{\Psi}_{\bm{\theta}_1}(\cdot)$ and the relation disentanglement controller $\bm{\Phi}_{\bm{\theta}_2}(\cdot)$ are trained jointly in the first phase. After obtaining the feature vectors $\{\bm{e}^{(0)}_n\}^{N_b}_{n=1}$, a prototypical classifier $\{\bm{w}_c\}^{|\bm{C}^{(0)}|}_{c=1}$ about base categories is utilized to draw the predictions $\{\bm{p}^{(0)}_n\in\mathbb{R}^{1\times 1\times |\bm{C}^{(0)}|}\}^{N_b}_{n=1}$ about the inputs by using Eq.~\ref{eq:3}
\begin{equation}
\bm{p}_{n,c}=\frac{\exp{(\frac{\bm{e}^{(0)}_{n} * \bm{w}_c}{||\bm{e}^{(0)}_{n}|| * ||\bm{w}_c||})}}{\sum_{m=1}^{|\bm{C}^{(0)}|} \exp{(\frac{\bm{e}^{(0)}_{n} * \bm{w}_m}{||\bm{e}^{(0)}_{n}|| * ||\bm{w}_m||})}}\text{.}
\label{eq:3}
\end{equation}

\noindent In the equation, $\bm{p}_{n,c}\in \bm{p}_{n}$ denotes the probability that the image $\bm{X}^{(0)}_n$ is classified as the base class $c$, and $\bm{w}_c$ represents the classifier weights about that category. The cross-entropy loss $\mathcal{L}_{ce}$ is utilized to train the backbone and the relation disentanglement controller to properly encode the embeddings of input images as well as optimize the prototypical classifier to correctly predict the labels of samples according to encoded embeddings
\begin{equation}
\mathcal{L}_{ce}=-\frac{1}{N_b}\sum_{n=1}^{N_b} \sum_{c=1}^{|\bm{C}^{(0)}|} \bm{y}_{n,c} *\log\bm{p}_{n,c}
\label{eq:4}
\end{equation}
\begin{equation}
s.t. \,\,\, \bm{y}_{n}=one\_hot\_encoding(y_{n})\text{,}\notag
\end{equation}

\noindent in which $one\_hot\_encoding(\cdot)$ denotes a one-hot encoding operation.

As mentioned before, it is difficult to directly disentangle spurious correlations between categories due to the poor controllability of training an FSCIL model. To overcome this challenge, an orthogonal proxy anchoring strategy is introduced in our method. In this strategy, a set of pre-defined orthogonal proxies $\bm{\varOmega}=\{\bm{\varsigma}_c\}^{|\bm{C}^{(0)}|}_{c=1}$ is adopted, where the proxy vectors obey the condition $\bm{\varsigma}_{c_1} * \bm{\varsigma}_{c_2}=0$ for $\forall c_1, c_2 \in \bm{C}^{(0)}$ if $c_1\neq c_2$. Also, an anchoring loss $\mathcal{L}_{ac}$ is employed, which is shown in Eq. \ref{eq:5} 
\begin{equation}
\mathcal{L}_{ac}=\sum_{c=1}^{|\bm{C}^{(0)}|} \|\bm{\alpha}_{c} - \bm{\varsigma}_c\|_2 \text{.}
\label{eq:5}
\end{equation}

\noindent In the above equation, $\bm{\alpha}_c\in\{\bm{\alpha}_c\}^{|\bm{C}^{(0)}|}_{c=1}$ represents the mean features of the base category $c$, while $\bm{\varsigma}_c$ denotes the orthogonal proxy about the class $c$. By minimizing the anchoring loss $\mathcal{L}_{ac}$, the representations of base categories can be controlled by orthogonal proxies. On one hand, the representations of base categories can be constrained into a fixed subspace. This is beneficial for avoiding feature aliasing between base and upcoming new categories, making it easier to disentangle spurious relation between the embeddings of base and novel categories. On the other hand, orthogonal proxies can ensure minimized interference between base categories and thus the accuracy on base class samples can be guaranteed. However, it is time-consuming and computationally expensive to iteratively calculate the mean features of base categories. Considering this issue, we utilize the weights of the prototypical classifier to approximate the mean features of base categories, \emph{i.e.}, $\bm{\alpha}_c\thickapprox \bm{w}_c$. The reason for the approximation is intuitive. According to Eq. \ref{eq:3} and Eq. \ref{eq:4}, the weights of the prototypical classifier are optimized to have maximized cosine similarity to feature embeddings that belong to the same class and thus they are constrained to be located in the center of embeddings per category.    

Meanwhile, a disentanglement proxy discriminability boosting strategy is
adopted to construct disentanglement proxies $\{\bm{\beta}_{n}\in\mathbb{R}^{1\times 1\times d'}\}^{N_d}_{n=1}$ for subsequently encountered novel categories, thus bridging gaps between different sessions. We hope that base categories should have low correlations to subsequently encountered categories. Moreover, the discriminability between the embeddings of novel categories should be boosted as much as possible. To realize this goal, a discriminability boosting loss $\mathcal{L}_{db}$ is employed in this strategy, which is given below:
\begin{align}
\mathcal{L}_{db}=&\underbrace{\sum_{n=1}^{N_d}\sum_{c=1}^{|\bm{C}^{(0)}|} \frac{\bm{\varsigma}_{c}*\bm{\beta}_{n}}{||\bm{\varsigma}_{c}||*||\bm{\beta}_{n}||}}_{\mathcal{L}_{base-novel}}\\
&\underbrace{+\sum_{n_1=1,n_1\neq n_2}^{N_d}\sum_{n_2=1}^{N_d} \frac{\bm{\beta}_{n_1}*\bm{\beta}_{n_2}}{||\bm{\beta}_{n_1}||*||\bm{\beta}_{n_2}||}}_{\mathcal{L}_{novel-novel}}\text{.}
\notag
\label{eq:6}         
\end{align}


\noindent The term $\mathcal{L}_{base-novel}$ aims to minimize the correlations between disentanglement proxies and base category embeddings, allowing disentanglement proxies to effectively guide the disentanglement for spurious relation between base and novel categories. In particular, as shown in Eq. 6, $\mathcal{L}_{base-novel}$ is not directly applied to the embeddings of base categories, as the constraint from $\mathcal{L}_{base-novel}$ may lower the quality of learned base class representations and make base class embeddings squeezed in feature space as shown in Fig. \ref{fig:h}. Instead, it minimizes the similarity between disentanglement proxies and pre-defined orthogonal proxies as base class embeddings are anchored by the orthogonal proxies in representation space. In contrast, the term $\mathcal{L}_{novel-novel}$ aims to boost the discriminability between disentanglement proxy vectors. Therefore, they can be discriminative enough to guide the disentanglement between spurious correlations of novel categories. Note after the training in each session, the mean features of encountered categories are extracted by the backbone network, which are used to constrain the drift of the relation disentanglement controller during incremental learning. The mean features of base categories extracted by the backbone are represented as $\{\bar{\bm{\alpha}}_c\}^{|\bm{C}^{(0)}|}_{c=1}$,

\subsubsection{Relation-disentanglement-guided adaptation}
In the second phase, the model is expanded to novel categories incrementally. Aiming to overcome the negative impact of data scarcity, the parameters of the backbone network are frozen in incremental sessions. For brevity, we take the $t$-$th$ session as an example to illustrate the second stage. Specifically, the backbone network $\bm{\Psi}_{\bm{\theta}_1}(\cdot)$ is first utilized to extract the visual embeddings $\big\{\bm{h}^{(t)}_n\big\}^{K|\bm{C}^{(t)}|}_{n=1}$ of given few-shot training samples $\big\{\bm{X}^{(t)}_n\big\}^{K|\bm{C}^{(t)}|}_{n=1}$. Then, the relation disentanglement controller $\bm{\Phi}_{\bm{\theta}_2}(\cdot)$ is utilized to further calibrate the extracted visual embeddings and obtain the features $\{\bm{e}^{(t)}_{n}\}^{K|\bm{C}^{(t)}|}_{n=1}$. Moreover, the mean class features extracted by the backbone network in previous ($t-1$) sessions $\{\bar{\bm{\alpha}}_c\}^{\sum^{t-1}_{j=0}|\bm{C}^{(j)}|}_{c=1}$ are also fed into the relation disentanglement controller, thereby producing the embeddings $\{\bm{\alpha}_c\}^{\sum^{t-1}_{j=0}|\bm{C}^{(j)}|}_{c=1}$. The above steps are as same as the procedures shown in Eq. \ref{eq:1} and Eq.~\ref{eq:2}. For simplicity, in this part, the embeddings of old categories and encountered novel ones are summarized into a unified formula, which are denoted as $\{\bar{\bm{e}}_{n}^{(t)}\}^{\sum^{t-1}_{j=0}|\bm{C}^{(j)}|+K|\bm{C}^{(t)}|}_{n=1}$ where the first $\sum^{t-1}_{j=0}|\bm{C}^{(j)}|$ elements denote the global embeddings of categories in previous ($t-1$) sessions while the last $K|\bm{C}^{(t)}|$ elements indicate the embeddings of the samples given in the $t$-$th$ session. Besides, the category labels of these embeddings are represented as $\{\bar{y}_{n}^{(t)}\}^{\sum^{t-1}_{j=0}|\bm{C}^{(j)}|+K|\bm{C}^{(t)}|}_{n=1}$.

Afterwards, aiming to guide the relation disentanglement controller $\bm{\Phi}_{\bm{\theta}_2}(\cdot)$ to disentangle spurious correlations between categories, a relation disentanglement strategy is employed, where a relation disentanglement loss $\mathcal{L}_{rd}$ is designed to properly rectify the relation disentanglement controller according to the guidance from the built disentanglement proxies. The details about the relation disentanglement loss $\mathcal{L}_{rd}$ are shown below: 
\begin{equation}
\mathcal{L}_{rd}= \sum^{N_b'}_{n=1} 
\begin{cases}
\gamma_{n} * (1-\frac{\bar{\bm{e}}_n^{(t)} * \bm{w}_{\bar{y}_n^{(t)}}}{||\bar{\bm{e}}_n^{(t)}||*||\bm{w}_{\bar{y}_n^{(t)}}||}) & \mbox{\textit{if}} \,\,\, \bar{y}_n^{(t)}\leq|\bm{C}^{(0)}| \\
\gamma_{n} * (1-\frac{\bar{\bm{e}}_n^{(t)} * \bm{\beta}_{\bar{y}_n^{(t)}-|\bm{C}^{(0)}|}}{||\bar{\bm{e}}_n^{(t)}||*||\bm{\beta}_{\bar{y}_n^{(t)}-|\bm{C}^{(0)}|}||}) & \mbox{\textit{else}} \,\,\, \\
\end{cases}
\label{eq:7}            
\end{equation}
 \begin{equation}
s.t. \,\, \gamma_{n}=\frac{N_{\bar{y}_n^{(t)}}'}{N_b'}\text{.}
\notag            
\end{equation}

\noindent In the equation, $\bar{y}_n^{(t)}$ represents the category label of the embeddings $\bar{\bm{e}}_{n}^{(t)}\in\{\bar{\bm{e}}_{n}^{(t)}\}_{n=1}^{N_b'}$. $N_{b}'$ denotes the batch size hyperparameter used in the second phase and $N_{\bar{y}_n^{(t)}}'$ represents the number of samples that belong to the category $\bar{y}_n^{(t)}$ in the mini-batch. Additionally, the balance weight $\gamma_n$ is utilized to balance the contribution of embeddings from different categories, thus reducing the bias brought by data imbalance and facilitating the training of the relation disentanglement controller. The relation disentanglement loss $\mathcal{L}_{rd}$ aims to align the embeddings of novel classes to disentanglement proxies, so as to boost the discriminability between the embeddings of encountered categories and guide the relation disentanglement controller to suppress spurious relation issues as much as possible.  Meanwhile, the mean embeddings of base categories are constrained to be aligned to the weights of the base class prototypical classifier, thereby reducing the drift of base class representations and ensuring they can be correctly categorized by the classifier. In this way, the spurious relation issue in FSCIL can be suppressed. After the training in the $t$-$th$ session, the weights of the base class prototypical classifier $\{\bm{w}_c\}^{|\bm{C}^{(0)}|}_{c=1}$ and the disentanglement proxies $\{\bm{\beta}\}_{n=1}^{\sum^{t}_{j=1}|\bm{C}^{(j)}|}\in\{\bm{\beta}_{n}\}^{N_d}_{n=1}$ are concentrated together to infer the categories of samples, where the inference is as same as the procedure given in Eq~\ref{eq:3}.

\begin{table*}
    \small
        \centering
        \caption{Experimental results on CIFAR-100. In the table, ``Mean'' denotes the mean accuracy of all sessions and ``Impr'' represents the improvement of our method with respect to the compared counterparts on mean accuracy. The best and the second-best results are highlighted by bold and underscores. Also, ``$\uparrow$'' or ``$\downarrow$'' represents that larger or smaller values indicate better performance. The terms used in Table II and Table III are as same as these in Table I, which are not introduced repeatedly.}
        \setlength{\tabcolsep}{2mm}{
        \begin{tabular}{c|c|ccccccccc|cl}
          \toprule
          & & \multicolumn{9}{c|}{\textbf{Accuracy in each session ($\%$)} $\uparrow$ } \\
          \cline{3-11}
          \textbf{Method} & \textbf{Ref} & 0 & 1 & 2 & 3 & 4 & 5 & 6 & 7 & 8 & \textbf{Mean} $\uparrow $ & \textbf{Impr} $\uparrow $\\
          \midrule
             iCaRL \cite{rebuffi2017icarl} & CVPR' 17 & 61.31 & 46.32 & 42.94 & 37.63 & 30.49 & 24.00 & 20.89 & 18.80 & 17.21 & 33.29 & +33.55\\
             EEIL \cite{castro2018end} & ECCV' 18 & 64.10 & 53.11 & 43.71 & 35.15 & 28.96 & 24.98 & 21.01 & 17.26 & 15.85 & 33.79 & +33.05 \\
             Rebalance \cite{hou2019learning} & CVPR' 19 & 64.10 & 53.05 & 43.96 & 36.97 & 31.61 & 26.73 & 21.23 & 16.78 & 13.54 & 34.22 & +32.62 \\
             TOPIC \cite{tao2020few} & CVPR' 20 & 64.10 & 55.88 & 47.07 & 45.16 & 40.11 & 36.38 & 33.96 & 31.55 & 29.37 & 42.62 & +24.22 \\
             CEC \cite{zhang2021few} & CVPR' 21 & 73.07 & 68.88 & 65.26 & 61.19 & 58.09 & 55.57 & 53.22 & 51.34 & 49.14 & 59.53 & +7.31 \\
             SPPP \cite{zhu2021self} & CVPR' 21 & 64.10 & 65.86 & 61.36 & 57.34 & 53.69 & 50.75 & 48.58 & 45.66 & 43.25 & 54.51 & +12.33 \\
             F2M \cite{shi2021overcoming} & NIPS' 21 & 64.71 & 61.99 & 58.99 & 55.58 & 52.55 & 49.96 & 48.08 & 46.28 & 44.67 & 53.65 & +13.19 \\
             LIMIT \cite{zhou2022few} & TPAMI' 22 & 73.81 & 72.09 & 67.87 & 63.89 & 60.70 & 57.77 & 55.67 & 53.52 & 51.23 & 61.83 & +5.01 \\
             FACT \cite{zhou2022forward} & CVPR' 22 & 74.60  & 72.09 & 67.56 & 63.52 & 61.38 & 58.36 & 56.28 & 54.24 & 52.10 & 62.24 & +4.60 \\
             MetaFSCIL \cite{chi2022metafscil} & CVPR' 22 & 74.50 & 70.10 & 66.84 & 62.77 & 59.48 & 56.52 & 54.36 & 52.56 & 49.97 & 60.79 & +6.05 \\
             C-FSCIL \cite{hersche2022constrained} & CVPR' 22 & 77.47 & 72.40 & 67.47 & 63.25 & 59.84 & 56.95 & 54.42 & 52.47 & 50.47 & 61.64 & +5.20 \\
             ALICE \cite{peng2022few} & ECCV' 22 & 79.00 & 70.50 & 67.10 & 63.40 & 61.20 & 59.20 & 58.10 & 56.30 & 54.10 & 63.21 & +3.63 \\
             CLOM \cite{zou2022margin} & NIPS' 22 & 74.20 & 69.83 & 66.17 & 62.39 & 59.26 & 56.48 & 54.36 & 52.16 & 50.25 & 60.57 & +6.27 \\
             ERDFR \cite{liu2022few} & ECCV' 22 & 74.40 & 70.20 & 66.54 & 62.51 & 59.71 & 56.58 & 54.52 & 52.39 & 50.14 & 60.78 & +6.06 \\
             MCNet \cite{ji2023memorizing} & TIP' 23 & 73.30 & 69.34 & 65.72 & 61.70 & 58.75 & 56.44 & 54.59 & 53.01 & 50.72 & 60.40 & +6.44 \\
             OSHHG \cite{cui2023rethinking} & TMM' 23 & 63.55 & 62.88 & 61.05 & 58.13 & 55.68 & 54.59 & 52.93 & 50.39 & 49.48 & 56.52 & +10.32 \\
             CABD \cite{zhao2023few} & CVPR' 23 & \underline{79.45} & \underline{75.38} & \underline{71.84} & \underline{67.95} & \underline{64.96} & \underline{61.95} & \underline{60.16} & \underline{57.67} & \textbf{55.88} & \underline{66.14} & +0.70 \\
             GKEAL \cite{zhuang2023gkeal} & CVPR' 23 & 74.01 & 70.45 & 67.01 & 63.08 & 60.01 & 57.30 & 55.50 & 53.39 & 51.40 & 61.35 & +5.49 \\
             DBONet \cite{guo2023decision} & CVPR' 23 & 77.81 & 73.62 & 71.04 & 66.29 & 63.52 & 61.01 & 58.37 & 56.89 & 55.78 & 64.93 & +1.91 \\
             SoftNet \cite{yoon2023soft} & ICLR' 23 & 72.62 & 67.31 & 63.05 & 59.39 & 56.00 & 53.23 & 51.06 & 48.83 & 46.63 & 57.57 & +9.27 \\
             WARP \cite{kim2023warping} & ICLR' 23 & 80.31 & 75.86 & 71.87 & 67.58 & 64.39 & 61.34 & 59.15 & 57.10 & 54.74 & 65.81 & +1.03 \\
             FCIL \cite{gu2023few} & ICCV' 23 & 77.12 & 72.42 & 68.31 & 64.47 & 61.18 & 58.17 & 56.06 & 54.19 & 52.02 & 62.66 & +4.18 \\
             S2C \cite{hu2023constructing} & arXiv' 23 & 75.15 & 73.07 & 68.31 & 64.61 & 61.94 & 59.41 & 57.62 & 55.62 & 53.19 & 63.21 & +3.63 \\
             TEEN \cite{wang2023few} & NIPS' 24 & 74.92 & 72.65 & 68.74 & 65.01 & 62.01 & 59.29 & 57.90 & 54.76 & 52.64 & 63.10 & +3.74 \\
             EHS \cite{deng2024expanding} & WACV' 24 & 71.27 & 67.40 & 63.87 & 60.40 & 57.84 & 55.09 & 53.10 & 51.45 & 49.43 & 58.87 & +7.97 \\
             EM \cite{zhu2024enhanced} & arXiv' 24 & 76.60 & 71.57 & 66.89 & 62.63 & 60.22 & 57.48 & 55.22 & 53.16 & 50.89 & 61.63 & +5.21 \\
             \midrule
             Our & - & \textbf{82.65} & \textbf{76.71} & \textbf{73.26} & \textbf{67.97} & \textbf{65.06} & \textbf{62.27} & \textbf{60.18} & \textbf{57.94} & \underline{55.54} & \textbf{66.84} & $-$ \\
            \bottomrule
        \end{tabular}}
        \label{tab:cifar}
    \end{table*}

\section{Experiments}
In this section, extensive experiments are conducted to validate the effectiveness of our CTRL-FSCIL method. We first introduce datasets and implementation details in Section-IV-A and Section-IV-B. Then, main experimental results are provided in Section-IV-C, and ablation study is conducted in Section-IV-D. 

\subsection{Datasets}
In this work, three widely used datasets are employed to evaluate the performance of CTRL-FSCIL, which are CIFAR-100 \cite{krizhevsky2009learning}, CUB-200 \cite{wah2011caltech}, and mini-ImageNet \cite{vinyals2016matching}. Specifically, CIFAR-100 and mini-ImageNet datasets both consist of 100 categories, and each category contains 500 training samples and 100 testing samples. The images in CIFAR-100 are in the resolution $32\times 32$, while mini-ImageNet contains samples with the size $84\times 84$. The CUB-200 dataset consists of 11788 images with the resolution $224\times 224$, collected from the birds of 200 categories, where training and testing samples are 5,994 and 5,794 respectively. Following the previous works \cite{tao2020few,zhang2021few,zhou2022forward}, for both CIFAR-100 and mini-ImageNet datasets, 60 categories are employed in the base session and the 5-way 5-shot problem is considered in each of 8 incremental sessions. Also, on CUB-200, the base session contains 100 categories and 10 incremental sessions all obey the 10-way 5-shot setting.

\subsection{Implementation Details}
For the experiments on CIFAR-100, mini-ImageNet, and CUB-200 datasets, the training strategies are slightly different. On the CIFAR-100 dataset, ResNet-12 \cite{he2016deep} is employed to build the backbone network of our CTRL-FSCIL. Also, we train the model for 100 epochs in the base session and 50$\thicksim$400 iterations in incremental sessions. The initial learning rate for the base session is set as 0.25, and the initial learning rate for incremental sessions are set as 0.2. The number $N_d$ of disentanglement proxies is set as 60. On the mini-ImageNet dataset, the backbone of CTRL-FSCIL is also built by using ResNet-12 and $N_d$ is set as 60. Moreover, we train our model for 80 epochs in the base session with the initial learning rate 0.25, and 75$\thicksim$200 iterations in incremental sessions with the initial learning rate 0.2. On the CUB-200 dataset, following \cite{peng2022few,zhou2022few}, we utilize ResNet-18 pretrained on ImageNet to implement the backbone of the model. Moreover, our model is trained for 45 epochs in the base session and 75$\thicksim$415 iterations in incremental sessions. The initial learning rates for base and incremental sessions are set as 0.025 and 0.02 respectively. The number $N_d$ of disentanglement proxies is set as 100. For all of these datasets, SGD is adopted to optimize learnable parameters, a two-layer MLP is employed to build the relation disentanglement controller, and the batch sizes $N_b$ and $N_b'$ for base and incremental sessions are set as 512 and 64 respectively.

\begin{table*}
    \small
        \centering
        \caption{Experimental results on the mini-ImageNet dataset. For the terms used in the table, please refer to the caption of Table I.}
        \setlength{\tabcolsep}{2mm}{
        \begin{tabular}{c|c|ccccccccc|cl}
            \toprule
            & & \multicolumn{9}{c|}{\textbf{Accuracy in each session ($\%$)} $\uparrow$ } \\
            \cline{3-11}
            \textbf{Method} & \textbf{Ref} & 0 & 1 & 2 & 3 & 4 & 5 & 6 & 7 & 8 & \textbf{Mean} $\uparrow $ & \textbf{Impr} $\uparrow $\\
            \midrule
            iCaRL  \cite{rebuffi2017icarl} & CVPR' 17 & 61.31 & 46.32 & 42.94 & 37.63 & 30.49 & 24.00 & 20.89 & 18.80 & 17.21 & 33.29 & +35.68 \\
            EEIL \cite{castro2018end} & ECCV' 18 & 61.31 & 46.58 & 44.00 & 37.29 & 33.14 & 27.12 & 24.10 & 21.57 & 19.58 & 34.97 & +34.00 \\
            Rebalance \cite{hou2019learning} & CVPR' 19 & 61.31 & 47.80 & 39.31 & 31.91 & 25.68 & 21.35 & 18.67 & 17.24 & 14.17 & 30.83 & +38.14 \\
            TOPIC \cite{tao2020few} & CVPR' 20 & 61.31 & 50.09 & 45.17 & 41.16 & 37.48 & 35.52 & 32.19 & 29.46 & 24.42 & 39.64 & +29.33 \\
            CEC \cite{zhang2021few} & CVPR' 21 & 72.00 & 66.83 & 62.97 & 59.43 & 56.70 & 53.73 & 51.19 & 49.24 & 47.63 & 57.75 & +11.22 \\
            F2M \cite{shi2021overcoming} & NIPS' 21 & 67.28 & 63.80 & 60.38 & 57.06 & 54.08 & 51.39 & 48.82 & 46.58 & 44.65 & 54.89 & +14.08 \\
            LIMIT \cite{zhou2022few} & TPAMI' 22 & 72.32 & 68.47 & 64.30 & 60.78 & 57.95 & 55.07 & 52.70 & 50.72 & 49.19 & 59.06 & +9.91 \\
            FACT \cite{zhou2022forward} & CVPR' 22 & 72.56 & 69.63 & 66.38 & 62.77 & 60.60 & 57.33 & 54.34 & 52.16 & 50.49 & 60.70 & +8.27 \\
            MetaFSCIL \cite{chi2022metafscil} & CVPR' 22 & 72.04 & 67.94 & 63.77 & 60.29 & 57.58 & 55.16 & 52.9 & 50.79 & 49.19 & 58.85 & +10.12 \\
            C-FSCIL \cite{hersche2022constrained} & CVPR' 22 & 76.40 & 71.14 & 66.46 & 63.29 & 60.42 & 57.46 & 54.78 & 53.11 & 51.41 & 61.61 & +7.36 \\
            ALICE \cite{peng2022few} & ECCV' 22 & 80.60 & 70.60 & 67.40 & 64.50 & 62.50 & 60.00 & 57.80 & 56.80 & 55.70 & 63.99 & +4.98 \\
            CLOM \cite{zou2022margin} & NIPS' 22 & 73.08 & 68.09 & 64.16 & 60.41 & 57.41 & 54.29 & 51.54 & 49.37 & 48.00 & 58.48 & +10.49 \\
            ERDFR \cite{liu2022few} & ECCV' 22 & 71.84 & 67.12 & 63.21 & 59.77 & 57.01 & 53.95 & 51.55 & 49.52 & 48.21 & 58.02 & +10.95 \\
            MCNet \cite{ji2023memorizing} & TIP' 23 & 72.33 & 67.70 & 63.50 & 60.34 & 57.59 & 54.70 & 52.13 & 50.41 & 49.08 & 58.64 & +10.33 \\
            OSHHG \cite{cui2023rethinking} & TMM' 23 & 60.65 & 59.00 & 56.59 & 54.78 & 53.02 & 50.73 & 48.46 & 47.34 & 46.75 & 53.04 & +15.93 \\
            CABD \cite{zhao2023few} & CVPR' 23 & 74.65 & 70.43 & 66.29 & 62.77 & 60.75 & 57.24 & 54.79 & 53.65 & 52.22 & 61.42 & +7.55 \\
            GKEAL \cite{zhuang2023gkeal} & CVPR' 23 & 73.59 & 68.90 & 65.33 & 62.29 & 59.39 & 56.70 & 54.20 & 52.59 & 51.31 & 60.48 & +8.49 \\
            DBONet \cite{guo2023decision} & CVPR' 23 & 74.53 & 71.55 & 68.57 & 65.72 & 63.08 & 60.64 & 57.83 & 55.21 & 53.82 & 63.44 & +5.53 \\
            SoftNet \cite{yoon2023soft} & ICLR' 23 & 77.17 & 70.32 & 66.15 & 62.55 & 59.48 & 56.46 & 53.71 & 51.68 & 50.24 & 60.86 & +8.11 \\
            WARP \cite{kim2023warping} & ICLR' 23 & 72.99 & 68.10 & 64.31 & 61.30 & 58.64 & 56.08 & 53.40 & 51.72 & 50.65 & 59.69 & +9.28 \\
            FCIL \cite{gu2023few} & ICCV' 23 & 76.34 & 71.40 & 67.10 & 64.08 & 61.30 & 58.51 & 55.72 & 54.08 & 52.76 & 62.37 & +6.60 \\
            S2C \cite{hu2023constructing} & arXiv' 23 & 73.25 & 71.57 & 67.46 & 64.01 & 61.04 & 58.41 & 55.62 & 53.62 & 52.00 & 61.89 & +7.08 \\
            TEEN \cite{wang2023few} & NIPS' 24 & 73.53 & 70.55 & 66.37 & 63.23 & 60.53 & 57.95 & 55.24 & 53.44 & 52.08 & 61.44 & +7.53 \\
            EHS \cite{deng2024expanding} & WACV' 24 & 69.43 & 64.86 & 61.30 & 58.21 & 55.49 & 52.77 & 50.22 & 48.61 & 47.67 & 56.51 & +12.46  \\
            EM \cite{zhu2024enhanced} & arXiv' 24 & \underline{81.28} & \underline{74.29} & \underline{70.07} & \underline{66.51} & \underline{63.80} & \underline{61.40} & \underline{57.99} & \underline{57.04} & \underline{56.53} & \underline{65.43} & +3.54 \\
            \midrule
            Our & - & \textbf{84.77} & \textbf{78.09} & \textbf{74.17} & \textbf{71.23} & \textbf{68.35} & \textbf{64.13} & \textbf{61.66} & \textbf{59.59} & \textbf{58.73} & \textbf{68.97} & $-$ \\
            \bottomrule
        \end{tabular}}
        \label{tab:mini}
    \end{table*}
  
    \begin{table*}
      \small
          \centering
          \caption{Experimental results on the CUB-200 dataset. For the terms used in the table, please refer to the caption of Table I.}
          \setlength{\tabcolsep}{1.5mm}{
          \begin{tabular}{c|c|ccccccccccc|cl}
              \toprule
              & & \multicolumn{11}{c|}{\textbf{Accuracy in each session ($\%$)} $\uparrow$ } \\
              \cline{3-13}
              \textbf{Method} & \textbf{Ref} & 0 & 1 & 2 & 3 & 4 & 5 & 6 & 7 & 8 & 9 & 10 &\textbf{Mean} $\uparrow $ & \textbf{Impr} $\uparrow $ \\
              \midrule
              iCaRL  \cite{rebuffi2017icarl} & CVPR' 17 & 68.68 & 52.65 & 48.61 & 44.16 & 36.62 & 29.52 & 27.83 & 26.26 & 24.01 & 23.89 & 21.16 & 36.67 & +30.99 \\
              EEIL \cite{castro2018end} & ECCV' 18 & 68.68 & 53.63 & 47.91 & 44.20 & 36.30 & 27.46 & 25.93 & 24.70 & 23.95 & 24.13 & 22.11 & 36.27 & +31.39 \\
              Rebalance \cite{hou2019learning} & CVPR' 19 & 68.68 & 57.12 & 44.21 & 28.78 & 26.71 & 25.66 & 24.62 & 21.52 & 20.12 & 20.06 & 19.87 & 32.49 & +35.17 \\
              TOPIC \cite{tao2020few} & CVPR' 20 & 68.68 & 62.49 & 54.81 & 49.99 & 45.25 & 41.40 & 38.35 & 35.36 & 32.22 & 28.31 & 26.28 & 43.92 & +23.74 \\
              CEC \cite{zhang2021few} & CVPR' 21 & 75.85 & 71.94 & 68.50 & 63.50 & 62.43 & 58.27 & 57.73 & 55.81 & 54.83 & 53.52 & 52.28 & 61.33 & +6.33 \\
              SPPP \cite{zhu2021self} & CVPR' 21 & 68.68 & 61.85 & 57.43 & 52.68 & 50.19 & 46.88 & 44.65 & 43.07 & 40.17 & 39.63 & 37.33 & 49.32 & +18.34 \\
              MOS \cite{cheraghian2021synthesized} & ICCV' 21 & 68.78 & 59.37 & 59.32 & 54.96 & 52.58 & 49.81 & 48.09 & 46.32 & 44.33 & 43.43 & 43.23 & 46.38 & +21.28 \\
              LIMIT \cite{zhou2022few} & TPAMI' 22 & 76.32 & 74.18 & 72.68 & 69.19 & \textbf{68.79} & 65.64 & 63.57 & 62.69 & 61.47 & 60.44 & 58.45 & 66.67 & +0.99 \\
              FACT \cite{zhou2022forward} & CVPR' 22 & 75.90 & 73.23 & 70.84 & 66.13 & 65.56 & 62.15 & 61.74 & 59.83 & 58.41 & 57.89 & 56.94 & 64.42 & +3.24 \\
              MetaFSCIL \cite{chi2022metafscil} & CVPR' 22 & 75.90 & 72.41 & 68.78 & 64.78 & 62.96 & 59.99 & 58.30 & 56.85 & 54.78 & 53.82 & 52.64 & 61.93 & +5.73 \\
              ALICE \cite{peng2022few} & ECCV' 22 & 77.40 & 72.70 & 70.60 & 67.20 & 65.90 & 63.40 & 62.90 & 61.90 & 60.50 & 60.60 & 60.10 & 65.75 & +1.91 \\
              ERDFR \cite{liu2022few} & ECCV' 22 & 75.90 & 72.14 & 68.64 & 63.76 & 62.58 & 59.11 & 57.82 & 55.89 & 54.92 & 53.58 & 52.39 & 61.52 & +6.14 \\
              CLOM \cite{zou2022margin} & NIPS' 22 & 79.57 & 76.07 & \underline{72.94} & \textbf{69.82} & \underline{67.80} & 65.56 & 63.94 & 62.59 & 60.62 & 60.34 & 59.58 & 67.17 & +0.49 \\
              MCNet \cite{ji2023memorizing} & TIP' 23 & 77.57 & 73.96 & 70.47 & 65.81 & 66.16 & 63.81 & 62.09 & 61.82 & 60.41 & 60.09 & 59.08 & 65.57 & +2.09 \\
              OSHHG \cite{cui2023rethinking} & TMM' 23 & 63.20 & 62.61 & 59.83 & 56.82 & 55.07 & 53.06 & 51.56 & 50.05 & 47.50 & 46.82 & 45.87 & 53.85 & +13.81 \\
              CABD \cite{zhao2023few} & CVPR' 23 & 79.12 & 75.37 & 72.80 & 69.05 & 67.53 & 65.12 & \underline{64.00} & \underline{63.51} & \textbf{61.87} & \underline{61.47} & \textbf{\textit{60.93}} & \underline{67.34} & +0.32 \\ 
              GKEAL \cite{zhuang2023gkeal} & CVPR' 23 & 78.88 & 75.62 & 72.32 & 68.62 & 67.23 & 64.26 & 62.98 & 61.89 & 60.20 & 59.21 & 58.67 & 66.35 & +1.31 \\
              DBONet \cite{guo2023decision} & CVPR' 23 & 78.66 & 75.53 & 72.72 & 69.45 & 67.21 & 65.15 & 63.03 & 61.77 & 59.77 & 59.01 & 57.42 & 66.34 & +1.32 \\
              SoftNet \cite{yoon2023soft} & ICLR' 23 & 78.11 & 74.51 & 71.14 & 62.27 & 65.14 & 62.27 & 60.77 & 59.03 & 57.13 & 56.77 & 56.28 & 63.95 & +3.71 \\
              WARP \cite{kim2023warping} & ICLR' 23 & 77.74 & 74.15 & 70.82 & 66.90 & 65.01 & 62.64 & 61.40 & 59.86 & 57.95 & 57.77 & 57.0 & 64.66 & +3.00 \\
              FCIL \cite{gu2023few} & ICCV' 23 & 78.70 & 75.12 & 70.10 & 66.26 & 66.51 & 64.01 & 62.69 & 61.00 & 60.36 & 59.45 & 58.48 & 65.70 & +1.96 \\
              S2C \cite{hu2023constructing} & arXiv' 23 & 75.92 & 73.57 & 71.67 & 68.01 & 66.94 & 63.61 & 62.22 & 61.42 & 59.79 & 58.56 & 57.46 & 65.38 & +2.28\\
              TEEN \cite{wang2023few} & NIPS' 24 & 77.26 & 76.13 & 72.81 & 68.16 & 67.77 & 64.40 & 63.25 & 62.29 & 61.19 & 60.32 & 59.31 & 66.63 & +1.03 \\
              EM \cite{zhu2024enhanced} & arXiv' 24 & \underline{80.83} & \underline{76.77} & 71.98 & 68.27 & 67.78 & \textbf{66.27} & 63.42 & 61.98 & 60.05 & 58.65 & 57.09 & 66.64 & +1.02 \\
              \midrule
              Our & - & \textbf{80.86} & \textbf{77.12} & \textbf{73.06} & \underline{69.53} & 67.33 & \underline{65.25} & \textbf{64.25} & \textbf{63.53} & \underline{61.68} & \textbf{61.51} & \underline{60.13} & \textbf{67.66} & $-$ \\
              \bottomrule
          \end{tabular}}
          \label{tab:cub}
      \end{table*}

\subsection{Main Results}

The performance of our CTRL-FSCIL on CIFAR-100, mini-ImageNet, and CUB-200 is given in Table \ref{tab:cifar}, Table \ref{tab:mini}, and Table \ref{tab:cub} respectively. In the following, we introduce these results in detail.

\begin{figure*}[t!]
\centering
\includegraphics[height=9cm]{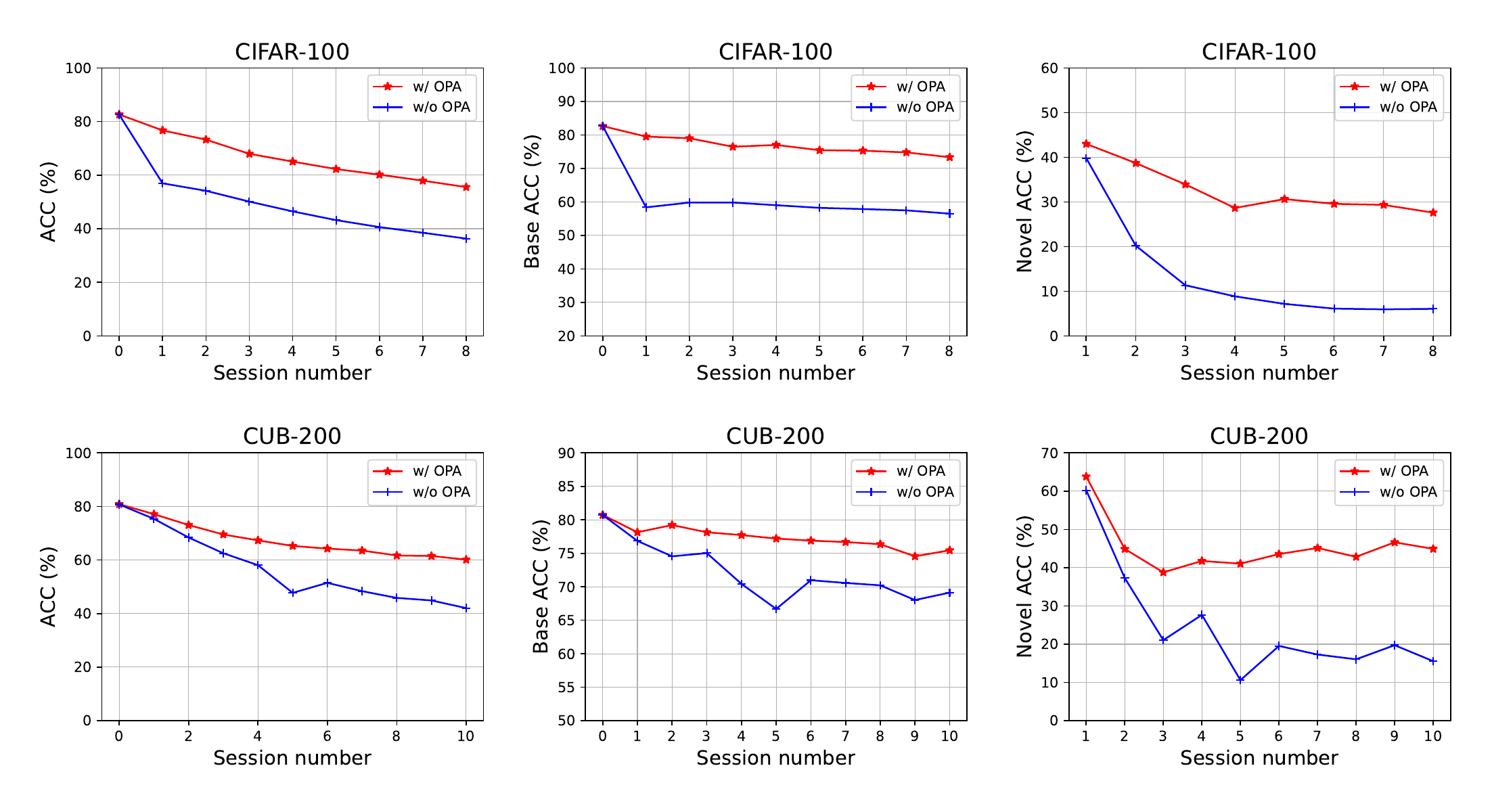}
\caption{The influencne of the Orthogonal Proxy Anchoring (OPA) strategy for the model. In the above figures, ``w/ OPA'' or ``w/o OPA'' indicates whether or not the OPA strategy is utilized in the first phase. ``Base Acc'', ``Novel Acc'', and ``Acc'' represent the accuracy on base categories, novel categories, and both of them respectively.}
\label{fig:d}
\end{figure*}

\begin{figure*}[t!]
\centering
\includegraphics[height=9cm]{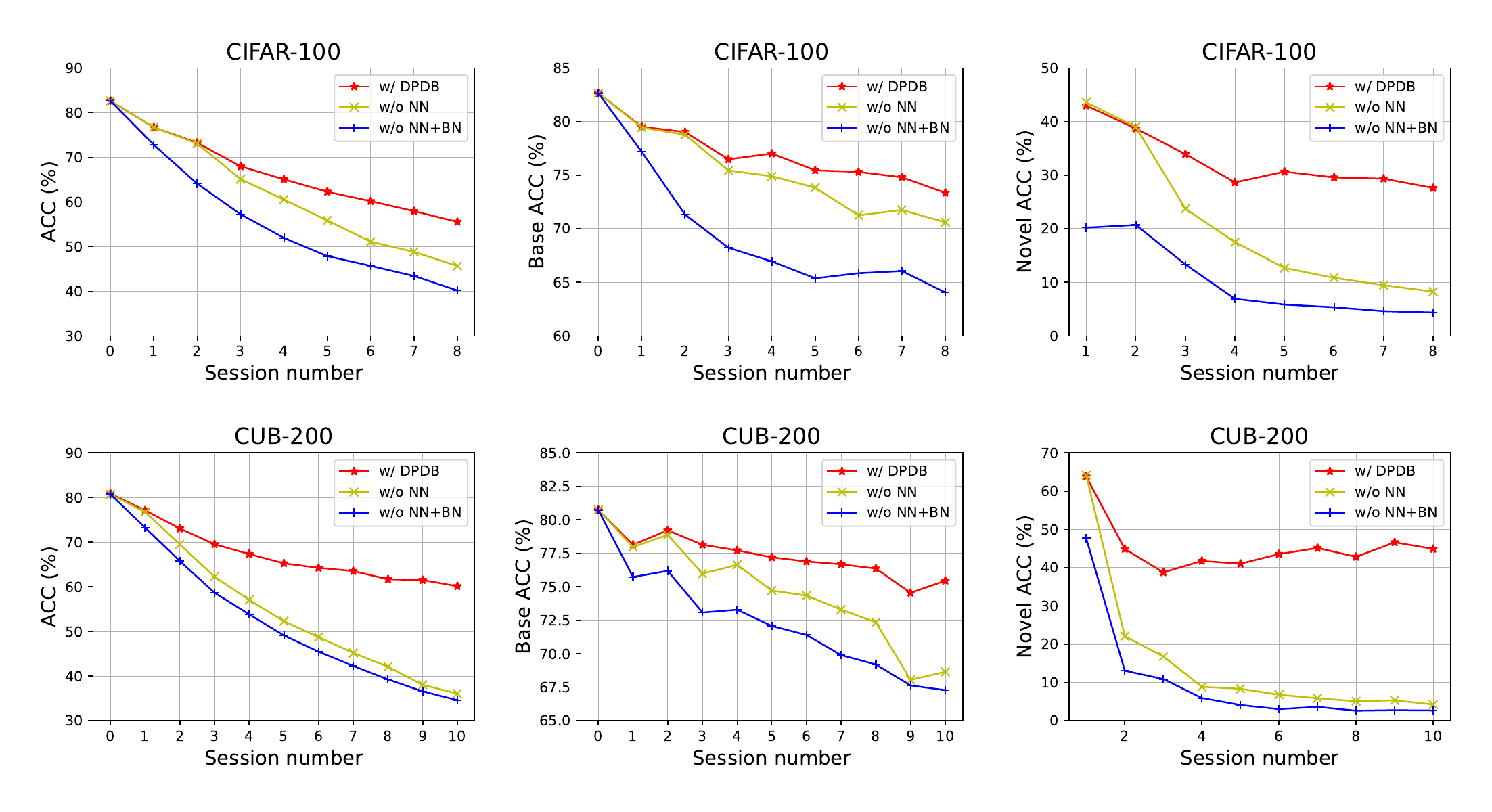}
\caption{The influencne of the Disentanglement Proxy Discriminability Boosting (DPDB) strategy for our method. In the above figures, ``w/ DPDB'' represents that the full DPDB strategy is utilized in the first phase. ``w/o NN'' indicates that the discriminability between disentanglement proxies is not ensured (\emph{i.e.}, $\mathcal{L}_{novel-novel}$ is removed from Eq. 6). ``w/o NN+BN'' indicates that the discriminability of disentanglement proxies to base categories is not considered further (\emph{i.e.}, $\mathcal{L}_{novel-novel}$ and $\mathcal{L}_{base-novel}$ are both removed).}
\label{fig:f}
\end{figure*}

\begin{figure*}[t!]
\centering
\includegraphics[height=4.5cm]{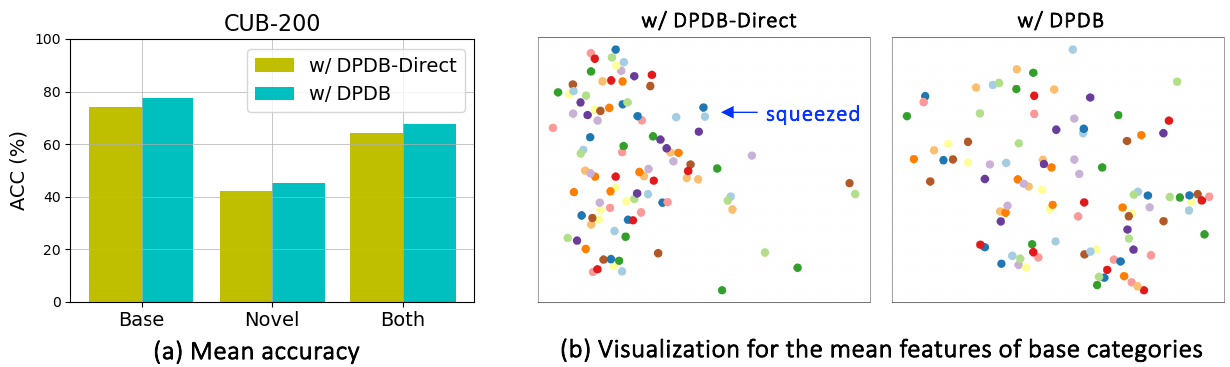}
\caption{The influence of directly applying the DPDB strategy to the embeddings of base categories. In ``w/ DPDB-Direct'', DPDB is applied to the representations of base categories directly, \emph{i.e.}, $\mathcal{L}_{base-novel}$ minimizes the similarity between base class embeddings and disentanglement proxies. In ``w/ DPDB'', $\mathcal{L}_{base-novel}$ is utilized to suppress the correlations between disentanglement proxies and pre-defined orthogonal proxies as base class embeddings are anchored by orthogonal proxies in feature space. In the tSNE visualization, each colored marker denotes the learned mean embeddings of a base category.}
\label{fig:h}
\end{figure*}

\begin{figure*}[t!]
\centering
\includegraphics[height=9cm]{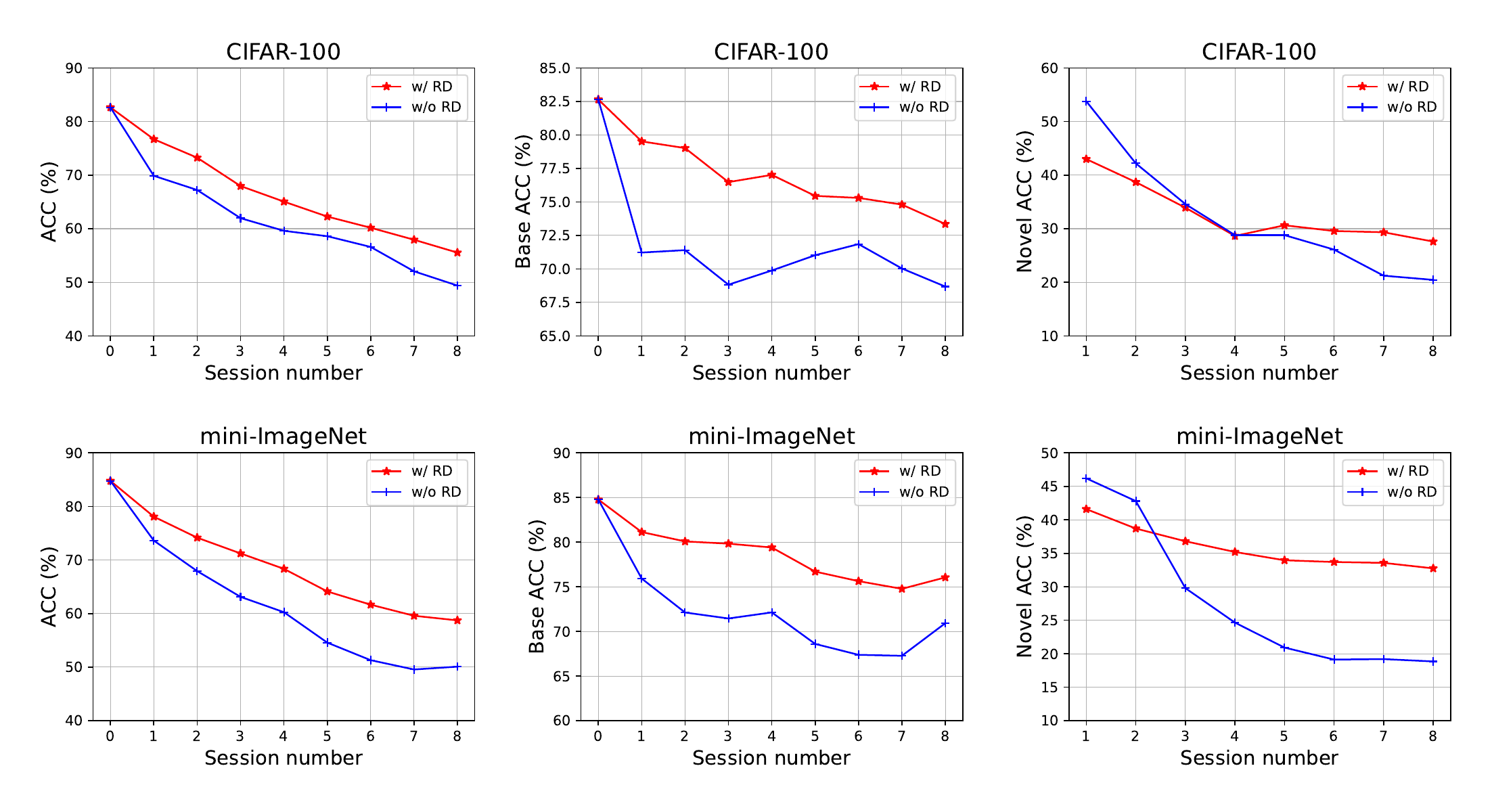}
\caption{The influence of the Relation Disentanglement (RD) for the model. In the figures, ``w/ RD'' indicates that the RD strategy is conducted to help adapt the model to encountered novel categories. ``w/o RD'' represents that the RD strategy is removed and our model is incrementally finetuned in each  session.}
\label{fig:e}
\end{figure*}

\subsubsection{CIFAR-100}
The results in Table \ref{tab:cifar} demonstrate that CTRL-FSCIL can achieve competitive performance on the CIFAR-100 dataset. On one hand, compared with the methods listed in the table, the accuracy of CTRL-FSCIL is obviously higher in most cases. For example, the accuracy of our CTRL-FSCIL is higher than that of the model FCIL \cite{gu2023few} by 5.53 \%, 3.29 \%, 4.95 \%, 3.50 \%, 3.88 \%, 4.10 \%, 4.12 \%, 3.75 \%, and 3.52 \% from the 0-$th$ session to the 8-$th$ session. Accordinglgy, the mean accuracy of FCIL \cite{gu2023few} is 4.18 \% lower that of our CTRL-FSCIL. On the other hand, the accuracy of our CTRL-FSCIL is slightly lower than that of CABD \cite{zhao2023few} in the last session of CIFAR-100. But in all of the other sessions, the performance of our CTRL-FSCIL is better than that of CABD \cite{zhao2023few}, \emph{e.g.}, the accuracy of CTRL-FSCIL is 1.33~\% and 1.42 \% higher than that of CABD \cite{zhao2023few} in the 1-th session and the 2-th session. 

\subsubsection{mini-ImageNet}
Similar results can also be found in mini-ImageNet. As can be seen from Table \ref{tab:mini}, the performance of CTRL-FSCIL is consistently better than that of the compared counterparts. For example, the accuracy of our CTRL-FSCIL is 3.49 \%, 3.80 \%, 4.10 \%, 4.72 \%, 4.55 \%, 2.73 \%, 3.67 \%, 2.55 \%, and 2.20 \% higher than that of EM \cite{zhu2024enhanced} in 9 sessions. In the meantime, the accuracy of CABD \cite{zhao2023few} is 10.12~\%, 7.66~\%, 7.88 \%, 8.46 \%, 7.60 \%, 6.89 \%, 6.87 \%, 5.94 \%, and 6.51 \% lower than that of CTRL-FSCIL. On this dataset, our CTRL-FSCIL also achieves the highest mean accuracy, which is higher than that of the second place \cite{zhu2024enhanced} by 3.54~\%. These results validate the competitive performance of our method again.

\subsubsection{CUB-200}
The experimental results on the CUB-200 dataset are summarized in Table \ref{tab:cub}. According to these results, we can draw the following conclusions. On one hand, our CTRL-FSCIL achieves the highest mean accuracy on the CUB-200 dataset, \emph{e.g.}, the mean accuracy of CTRL-FSCIL is 0.32 \% higher than that of the scend place \cite{zhao2023few}, and 0.49~\% higher than that of the third place \cite{zou2022margin}. On the other hand, the performance of our CTRL-FSCIL in each session is competitive as well. For example, among the total 11 sessions, the performance of CTRL-FSCIL ranks the first place in the 6 sessions and the second place in the 4 sessions. Although neither the first place nor the second place is achieved by our method in the 4-$th$ session, the accuracy of CTRL-FSCIL is only 1.46 \% and 0.47 \% lower than that of LIMIT \cite{zhou2022few} and CLOM \cite{zou2022margin}.

\begin{figure*}[t!]
\centering
\includegraphics[height=4.5cm]{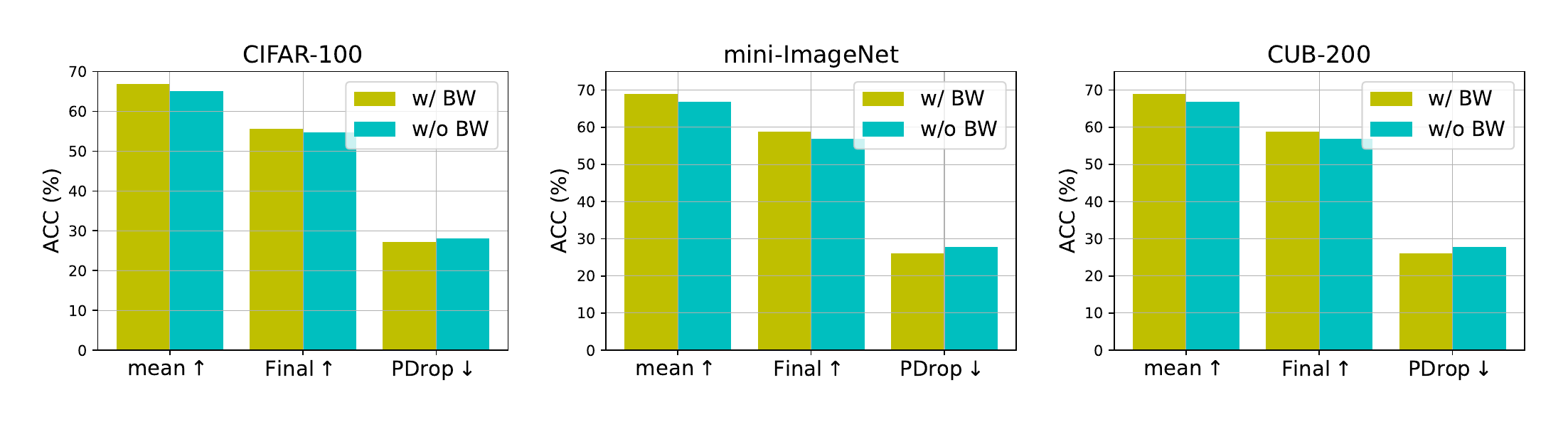}
\caption{The influence of Balance Weights (BW) for accuracy. In the above figures, ``w/ BW'' or ``w/o BW'' indicates whether or not balance weights are employed in Eq. \ref{eq:7}. Besides, ``Final'' indicates the accuracy achieved in the last session, ``Mean'' represents the mean accuracy of all sessions, and ``PDrop'' denotes the accuracy drop between the last and the first session. In particular, ``$\uparrow$'' or ``$\downarrow$'' represents that larger or smaller values indicate better performance.}
\label{fig:g}
\end{figure*}

\begin{figure*}[t!]
\centering
\includegraphics[height=4cm]{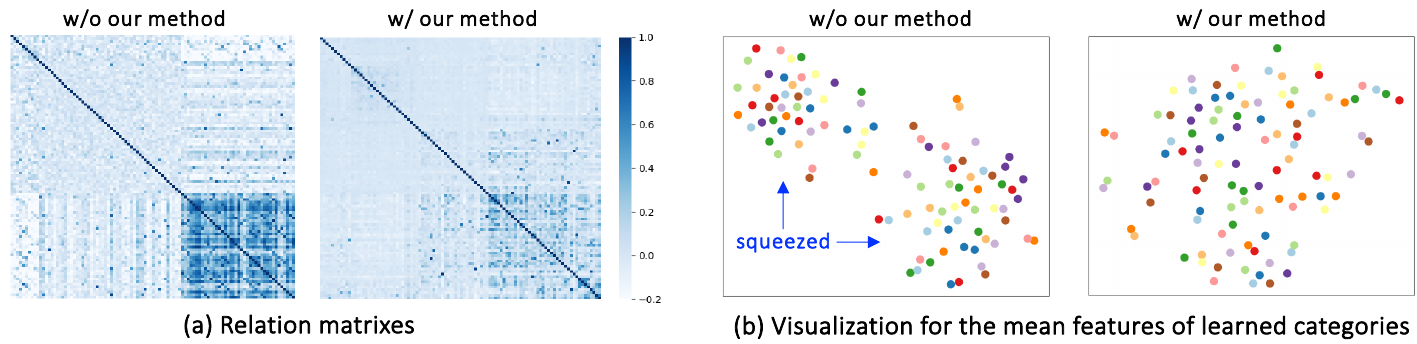}
\caption{Visualized analysis for the influence of our method on suppressing spurious correlations between categories, which is conducted in the last session of mini-ImageNet. The relationships between categories learned by our method is provided in (a), and the incrementally learned mean embeddings of categories are visualized in (b).}
\label{fig:b}
\end{figure*}

\subsection{Method analysis}
In this subsection, we provide the analysis for each component of our CTRL-FSCIL method. We first study the influence of the Orthogonal Proxy Anchoring (OPA) strategy in Fig.~\ref{fig:d}. The experimental results in the figure indicate that the use of the OPA strategy (``w/~OPA'') can obviously help the model to achieve better performance. However, when the OPA strtategy is removed (``w/o OPA'') from the first phase, the performance of our model on base and novel categories drops substaionally in incremental sessions, as the removel of OPA makes it difficult to control the representations of base categories and thereby intensifies the interference between base and novel categories. Similarly, according to the results in Fig. \ref{fig:f}, we can draw the following conclusions for the Disentanglement Proxy Discriminability Boosting (DPDB) strategy. On one hand, the use of the full DPDB strategy (``w/~DPDB'') can guarantee the discriminability of disentanglement proxies and minimize their correlations to the embeddings of base categories. Therefore, spurious relation between categories can be suppressed by considering the guidance of disentanglement proxies, making the performance of the model more accurate. On the other hand, however, when the discriminability between disentanglement proxy vectors is not ensured (``w/o NN''), the performance of our model gets an obvious decrease on both base and novel categories, as relation aliasing between novel categories misleads the update of the relation disentanglement controller and results in the drift of old knowledge. In addition, when the discriminability of disentanglement proxies to base categories is further not considered (``w/o NN+BN''), inter-class interference between base and novel categories further degrades the performance of our model. These results demonstrate the important role of constructing discriminative disentanglement proxies during disentangling spurious relation between categories. Furthermore, in Fig. \ref{fig:h}, we also investigate the influence of directly applying the DPDB strategy to the representations of base categories. According to Fig. \ref{fig:h} (b), we can see that directly applying the DPDB strategy to the representations of base categories (``w/~DPDB-Direct'') has negative impact on learning high-quality embeddings of base categories, \emph{e.g.}, the global embeddings of base categories are squeezed in feature space as shown in the figure. In contrast, our DPDB strategy (``w/ DPDB'') can obviously learn better base class representations, \emph{e.g.}, the global embeddings of base categories are more uniformly distributed in feature space, which are more discriminative to reduce interference between categories. Correspondingly, the results in Fig. \ref{fig:h} (a) also indicate that the performance of ``w/ DPDB'' is better than that of ``w/~DPDB-Direct''.

In Fig. \ref{fig:e}, we study the influence of the Relation Disentanglement (RD) strategy. The experimental results demonstrate the effectiveness of our method again. The results in Fig.~\ref{fig:e} indicate that when removing the RD strategy (``w/o~RD''), the performance of our model drops substaionally in incremental sessions due to the influence of spurious correlations between categories. In constract, the RD strategy (``w/ RD'') can obviously help the model to achieve better performance on base and novel categories. During the second phase, balance weights are utilized in Eq. \ref{eq:7} to balance the contribution of embeddings from different categories and guide the model to treat each category equally. In Fig \ref{fig:g}, we quantitatively analyze the influence of these balance weights. The results in the figure indicate that the use of balance weghts are beneficial for training a better relation disentanglement controller and making the performance of the model more accurate. Finally, we qualitatively analyze the influence of our method on alleviating spurious relation issues in Fig \ref{fig:b}. Firstly, we visualize the category relation learned by using or not using our approach in Fig \ref{fig:b}~(a). The results indicate that our method can effectively suppress spurious correlations between categories. In Fig \ref{fig:b}~(b), we also visualize the gobal embeddings of categories incrementally learned in each session. On one hand, the embeddings learned by our method are more discriminatively distributed in representation space, which are more beneficial for suppressing interference between categories during label prediction. On the other hand, without using our proposed method, the representations of incrementally learned categories are squeezed in feature space, which raises the risk of interference between categories and makes the model performance unsatisfactory. These results demonstrate the effectiveness of our approach.

\section{Conclusion}
In this paper, we propose a new simple-yet-effective method, called \textbf{C}on\textbf{T}rollable \textbf{R}elation-disentang\textbf{L}ed \textbf{F}ew-\textbf{S}hot \textbf{C}lass-\textbf{I}ncremental \textbf{L}earning (CTRL-FSCIL), which tackles FSCIL from a new perspective, \emph{i.e.}, disentangling spurious relation between categories. Specifically, in the base session, our method anchors base category representations in feature space and constructs disentanglement proxies to bridge gaps between embeddings of different sessions, so as to make category relation controllable. In incremental sessions, a disentanglement loss is designed to effectively guide a relation disentanglement controller to disentangle spurious correlations between the features encoded by the backbone. In this way, spurious relation issues in FSCIL can be suppressed. Extensive experiments on CIFAR-100, mini-ImageNet, and CUB-200 datasets validate the effectiveness of our CTRL-FSCIL. In the future, we plan to further extend our method to other relevant tasks, such as incremental few-shot semantic segmentation and object detection.

\bibliographystyle{IEEEtran}
\bibliography{ref}

\begin{IEEEbiography}[{\includegraphics
[width=1in,height=1.25in,clip,
keepaspectratio]{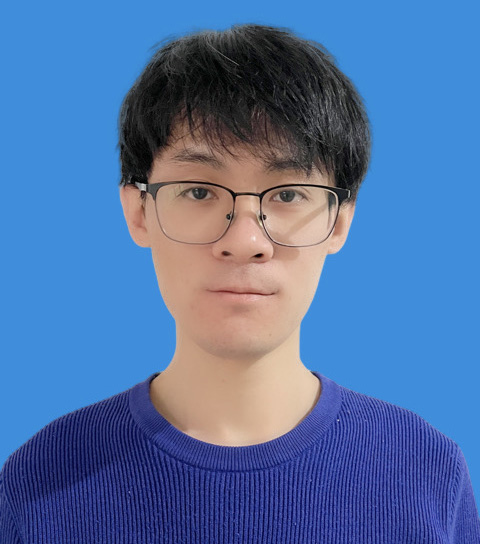}}]
{Yuan Zhou}
received Ph.D Degree from Hefei University of Technology. He was with Key Laboratory of Knowledge Engineering with Big Data (Hefei University of technology), Ministry of Education. His research interests include few-shot learning and image segmentation.
\end{IEEEbiography}

\begin{IEEEbiography}[{\includegraphics
[width=1in,height=1.25in,clip,
keepaspectratio]{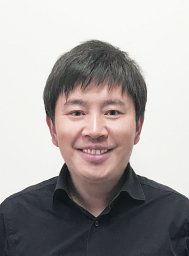}}]
{Richang Hong}
received the Ph.D. degree from the University of Science and Technology of China, Hefei, China, in 2008. He was a Research Fellow of the School of Computing with the National University of Singapore, from 2008 to 2010. He is currently a Professor with the Hefei University of Technology, Hefei. He is also with Key Laboratory of Knowledge Engineering with Big Data (Hefei University of technology), Ministry of Education. He has coauthored over 100 publications in the areas of his research interests, which include multimedia content analysis and social media. He is a member of the ACM and the Executive Committee Member of the ACM SIGMM China Chapter. He was a recipient of the Best Paper Award from the ACM Multimedia 2010, the Best Paper Award from the ACM ICMR 2015, and the Honorable Mention of the IEEE Transactions on Multimedia Best Paper Award. He has served as the Technical Program Chair of the MMM 2016, ICIMCS 2017, and PCM 2018. Currently, he is an Associate Editor of IEEE Transactions on Big Data, IEEE Transactions on Computational Social System, ACM Transactions on Multimedia Computing Communications and Applications, Information Sciences (Elsevier), Neural Processing Letter (Springer) and Signal Processing (Elsevier).
\end{IEEEbiography}

\begin{IEEEbiography}[{\includegraphics
[width=1in,height=1.25in,clip,
keepaspectratio]{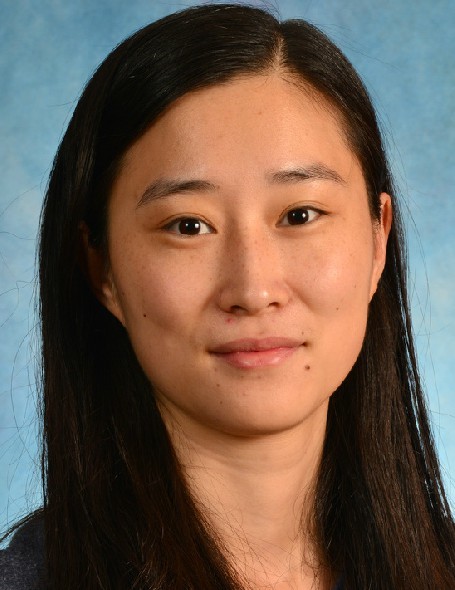}}]
{Yanrong Guo}
is a professor at School of Computer and Information, Hefei University of Technology (HFUT). She is also with Key Laboratory of Knowledge Engineering with Big Data (Hefei University of technology), Ministry of Education. She received her Ph.D. degree at HFUT in 2013. She was a postdoc researcher at University of North Carolina at Chapel Hill (UNC) from 2013 to 2016. Her research interests include image analysis and pattern recognition.
\end{IEEEbiography}

\begin{IEEEbiography}[{\includegraphics
[width=1in,height=1.25in,clip,
keepaspectratio]{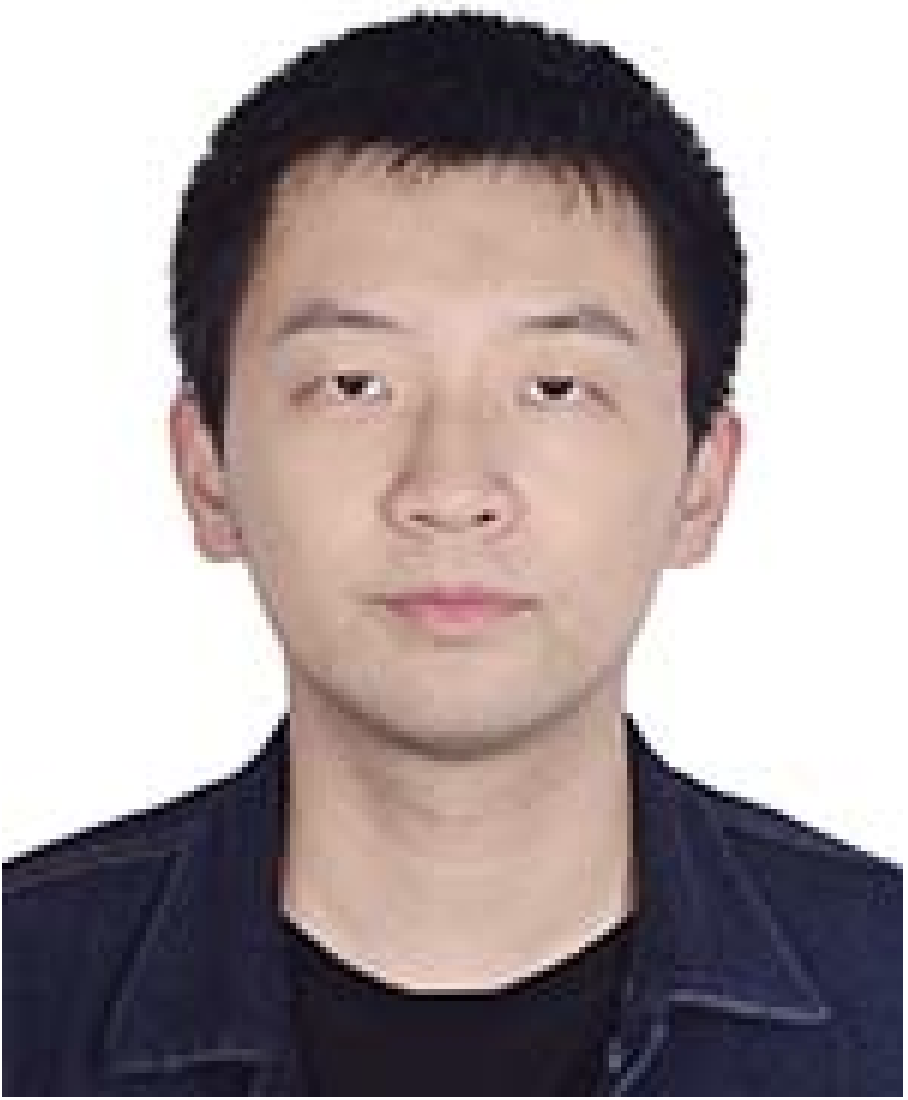}}]
{Lin Liu}
received the B.S. degree from University of Science and Technology of China, Hefei, China in 2019. He is currently pursuing the PhD degree with University of Science and Technology of China. His research interests include computer vision and machine learning.
\end{IEEEbiography}

\begin{IEEEbiography}[{\includegraphics
[width=1in,height=1.25in,clip,
keepaspectratio]{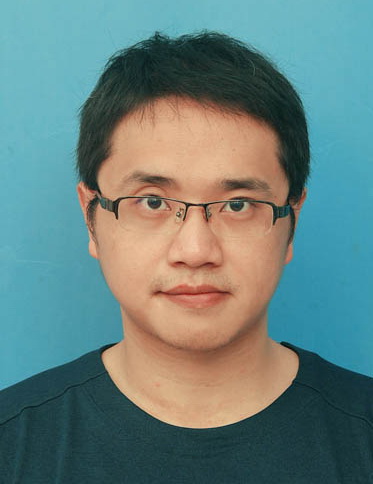}}]
{Shijie Hao}
is a professor at School of Computer Science and Information Engineering, Hefei University of Technology (HFUT). He is also with Key Laboratory of Knowledge Engineering with Big Data (Hefei University of technology), Ministry of Education. He received his Ph.D. degree at HFUT in 2012. His research interests include image processing and multimedia content analysis.
\end{IEEEbiography}

\begin{IEEEbiography}[{\includegraphics
[width=1in,height=1.25in,clip,
keepaspectratio]{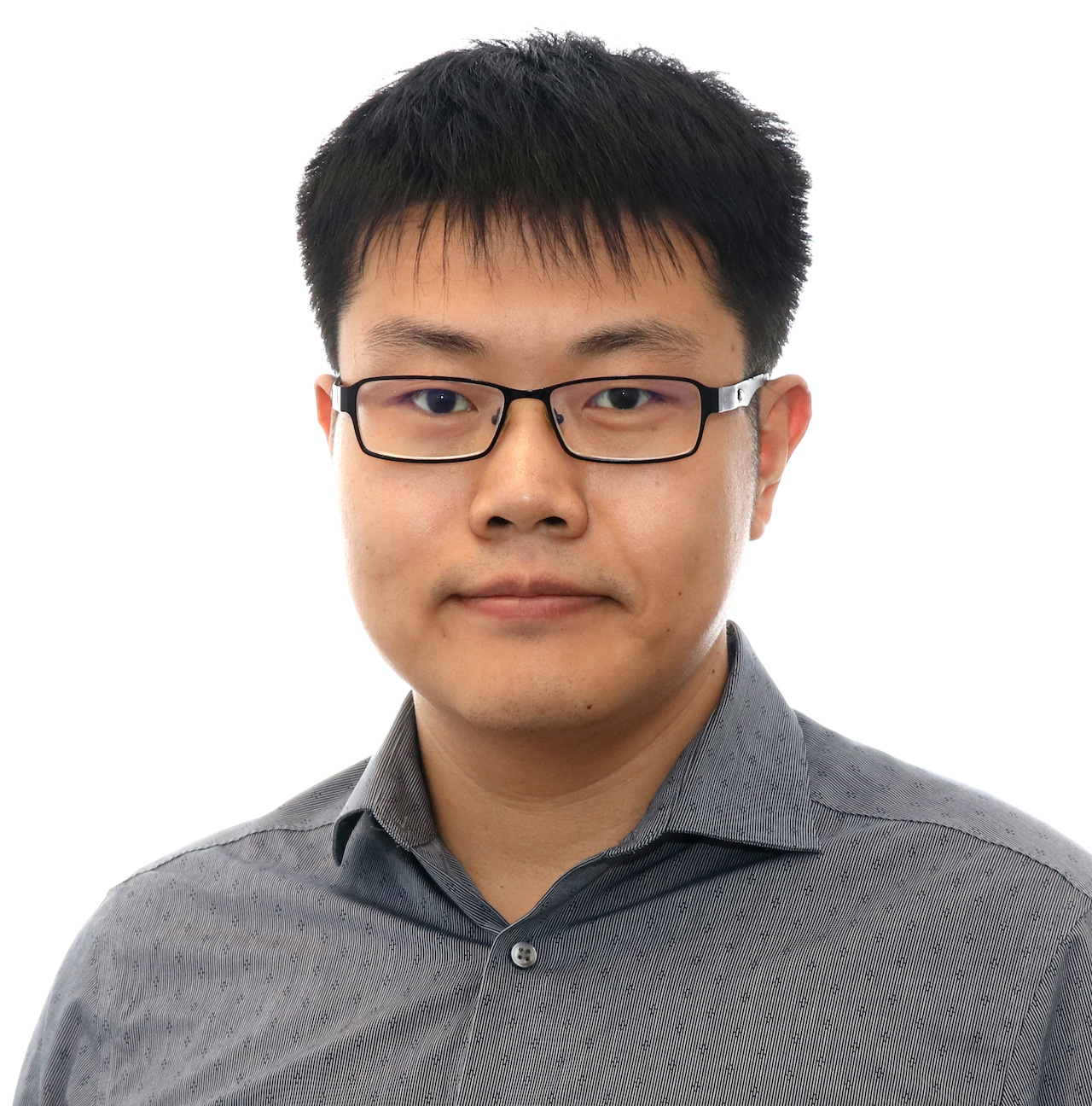}}]
{Hanwang Zhang}
is currently an associate professor at Nanyang Technological University, Singapore. He was a research scientist at the Department of Computer Science, Columbia University, USA. He has received the B.Eng (Hons.) degree in computer science from Zhejiang University, Hangzhou, China, in 2009, and the Ph.D. degree in computer science from the National University of Singapore in 2014. His research interest includes computer vision, multimedia, and social media. He is the recipient of the Best Demo runner-up award in ACM MM 2012, the Best Student Paper award in ACM MM 2013, and the Best Paper Honorable Mention in ACM SIGIR 2016, and TOMM best paper award 2018. He is also the winner of Best Ph.D. Thesis Award of School of Computing, National University of Singapore, 2014.
\end{IEEEbiography}

\end{document}